\documentclass{article}

\usepackage[final]{neurips_2021}
\usepackage[utf8]{inputenc}
\usepackage[T1]{fontenc}
\usepackage{hyperref}
\usepackage{url}
\usepackage{booktabs}
\usepackage{amsmath}
\usepackage{amsfonts}
\usepackage{nicefrac}
\usepackage{microtype}
\usepackage{xcolor}
\usepackage{graphicx}
\usepackage{subcaption}

\title{Disaster mapping from satellites: damage detection with crowdsourced point labels}

\author{
    Danil~Kuzin\\
    Department of Computer Science\\
    Department of Physics\\
    Lancaster University, Lancaster, UK\\
    \texttt{d.kuzin@lancaster.ac.uk} \\
  \And
  Olga~Isupova\\
    Department of Computer Science\\
    University of Bath, Bath, UK\\
    \texttt{oi260@bath.ac.uk}
  \And
  Brooke~D.~Simmons\\
    Department of Physics\\
    Lancaster University, Lancaster, UK\\
    \texttt{b.simmons@lancaster.ac.uk}
  \And
  Steven~Reece\\
    Department of Engineering Science\\
    University of Oxford, Oxford, UK\\
    \texttt{reece@robots.ox.ac.uk}
}

\begin{document}

\maketitle

\begin{abstract}
  High-resolution satellite imagery available immediately after disaster events is crucial for response planning as it facilitates broad situational awareness of critical infrastructure status such as building damage, flooding, and obstructions to access routes.  Damage mapping at this scale would require hundreds of expert person-hours.  However, a combination of crowdsourcing and recent advances in deep learning reduces the effort needed to just a few hours in real time.  Asking volunteers to place point marks, as opposed to shapes of actual damaged areas, significantly decreases the required analysis time for response during the disaster.  However, different volunteers may be inconsistent in their marking.  This work presents methods for aggregating potentially inconsistent damage marks to train a neural network damage detector.
\end{abstract}

\section{Introduction}\label{sec:introduction}
  Maps of damage and the health of infrastructure are key urgent needs of responders and decision makers in the immediate aftermath of a crisis.  Satellite imagery is useful for datasets mapping as it can cover large areas rapidly and uniformly at high enough resolution to identify major damage.  Machine learning approaches promise a rapid assessment of arbitrarily large amounts of satellite data via assessment of changes between pre-~and post-event images, but disaster mapping in particular has many challenges, including weather conditions, low resolution of images, diversity of environments, variable temporal intervals between collection of images, difference in projections due to movement of satellites.  These factors pose major obstacles to out of the box machine learning algorithms  and existence of universally applicable datasets. 

  Typical approaches to damaged building detection from satellite imagery consist of two stages: localization of building footprints and classification of the damage severity for each of these footprints based on pre-~and post-event imagery.  The current state-of-the-art approaches for both of these stages use neural networks of various architectures with different approaches to the combination of pre-~and post-event images.  The localization of building footprints can be viewed as an object detection problem of predicting the bounding box for each building; or as an instance segmentation problem of predicting the pixel mask for each building.  Both object detection and segmentation approaches have been used for building footprint localization including SSD~\citep{li2019building}, R-CNN~\citep{weber2020building}, and LinkNet~\citep{golovanov2018building}.  Damage detection approaches combine pre-~and post-event imagery either on the neural network input layer by stacking together both optical images into a 6-channel image~\citep{xu2019building}, or by processing each image via a separate feature extraction neural network~\citep{weber2020building} and concatenating features before performing detection.  These were succeeded by cross region transfer learning~\citep{xu2019building} and simultaneous segmentation and classification with dilated residual networks~\citep{gupta2020rescuenet}.

  All the above models were trained on large, carefully-prepared, high-resolution datasets.  Some of these datasets are available online and include building footprint datasets and damaged building datasets.  However, careful data preparation is not always possible during time-critical disaster responses.  Therefore, we focus our attention on mapping damage with free resources and in a timely manner, the crucial factors for a disaster response operation. This involves images captured under non-ideal conditions, at varying resolutions, and labeled with volunteer {\it point marks}.  Point marks are single dot-like labels that we ask our volunteers to place on satellite imagery.  They are much faster to collect than bounding boxes or segmented building footprints, and the speed is highly desirable during a live response deployment. These marks must then be aggregated to identify a consensus for each building as different volunteers may have different opinions about the same object.  Our work is two-fold: we first create a training dataset of damage severity on building footprints using only the point marks from volunteers, and then demonstrate how this dataset can be used to train a neural network to map damages on new unseen data. The second step allows us to cover more areas to provide information for disaster responders without the need to rely on volunteers.

\subsection{Application Context}\label{sec:application_context}
  In 2017 hurricanes Irma and Maria impacted multiple islands in the Caribbean.  Disaster imagery was made available from multiple sources including NASA's Landsat 8, ESA's Sentinel-2, Planet's Dove~(\url{planet.com}), RapidEye and SkySat constellations, and Maxar's~(formerly DigitalGlobe) satellite constellation~(\url{maxar.com}).  We focus here on data from live deployments by the Planetary Response Network~(PRN, \url{planetaryresponsenetwork.org}), a collaboration between Zooniverse crowdsourcing platform~(\url{zooniverse.org}), response and resilience organizations, and machine learning researchers.  We present a data processing pipeline for detecting damaged buildings, which we developed primarily using imagery and labels from PRN deployments in the Caribbean.
  
  Previous work has addressed the identification of general damage levels in individual satellite image segments. Mapping individual building damage was requested by Rescue Global, a humanitarian organisation with which we regularly deploy following disasters. Motivated by these needs, we focus here on identifying damage to individual structures. This need is common in the humanitarian assistance and disaster response (HADR) domain as it directly feeds ongoing situational awareness for stakeholders.
  
  Automated building damage detection from satellite imagery is a significant current research problem:
  \citet{shen2020cross} fuse pre-~and post-event images for damage detection, \citet{lee2020assessing} use semi-supervised learning to reduce the amount of training data required for damage detection, \citet{benson2020assessing} assess generalization error of damage detection models on new datasets, \citet{boin2020multi} address the problem of class imbalance for damaged building detection, \citet{xu2019building} discuss different architectures for building damage assessment.

  In this context our work aims to reduce the required time for mapping of training data during the disasters in new conditions, where the existing models need to be fine-tuned.  This is achieved by using crowdsourced marks for labelling. Crowdsourcing has proven to be useful during our live deployments before when we carefully aggregated crowd labels to remove any inconsistencies between volunteers.

\section{The Crowd-labelled Dataset}\label{sec:dataset}
  Our aim is to detect objects of an arbitrary shape, specifically building footprints, from marks that only point to the objects.  In this section we describe how we create the dataset from the pre-~and post-event satellite images and marked points of damage by the crowd.  First, we detect building footprints on both images, using a segmentation neural network, trained on a building footprints dataset.  After that, we extract bounding boxes for segmented buildings and aggregate marks of the crowd for every footprint.

\subsection{Building Footprint Detection}\label{sec:dataset:building_footprint_detection}
  There are several approaches for point supervision that we can apply to convert point marks into damaged building footprints.  \citet{bearman2016s} use the pre-trained objectness prior that is trained on other datasets; it assigns probabilities that pixels belong to objects and these probabilities are included in the loss function.  \citet{papadopoulos2017training} use points in the centers of the objects and estimate object sizes based on them.  Points on the borders of the objects can be used to define the proposal object masks: \citet{maninis2018deep} use extreme points for the objects, generate Gaussian probabilistic masks around the points, bound by an exterior of marked points and then generate masks for the objects to use as additional layers in a network; \citet{zhou2019bottom} use CornerNet and HourglassNet to detect corner points and center points for objects of each class, then group extreme points associated with centers, aggregate edges for extreme points, and extract octagon masks.  \citet{benenson2019large} use corrective marks to specify areas that do or do not belong to objects, and create binary disks around these marks.

  Instead of using marks specifically to determine shapes of objects, we ask volunteers to provide marks anywhere on damaged buildings to speed up the labelling process. In our case building footprints can be used for limiting the area of interest around these point marks.  There exist multiple datasets to train a method for building footprint localisation, such as SpaceNet~(\url{spacenet.ai}), Open Cities AI Challenge~(\url{drivendata.org/competitions/60/building-segmentation-disaster-resilience}), DeepGlobe2018~\citep{demir2018deepglobe} and damaged building datasets such as xView~\citep{lam2018xview} and xBD~\citep{gupta2019xbd}.

  We use the xView2 dataset to train the ensemble of UNet meta-architectures~\citep{ronneberger2015u}, that was used in the building localization solution for the xView2 challenge~(\url{https://github.com/DIUx-xView/xView2_first_place}).  The neural networks are used separately for pre-~and post-event images to detect building footprints.

  Satellite images are often obtained with different off-nadir angles at different days and therefore they do not align well for precise per-pixel analysis.  To mitigate this problem we associate buildings between pre-~and post-event images using their bounding boxes instead of segmentation masks.  Segmentation masks are thresholded with the optimised value, then contours are extracted from binarised masks, converted to polygons, and their boundaries are used as bounding boxes. This approach is less sensitive to imagery misalignments and pixel-level precision is not usually required for disaster responders.

  We evaluate the prediction quality on the expertly labelled subset of images with bounding boxes, results are given in Table~\ref{table:prediction_quality_of_all_bounding_boxes}.  Average precision (AP\textsuperscript{IoU}) metric details are given in common objects in context (COCO) challenges~(\url{https://cocodataset.org/\#detection-eval}), precision and recall in the context of object detection are defined as in PASCAL VOC challenges~(\url{http://host.robots.ox.ac.uk/pascal/VOC/pubs/everingham15.pdf}).  As it can be seen our ensemble method achieve reasonable results on both types of images with the higher performance on the pre-event images. The neural networks are trained on undamaged buildings, whereas the post-event images capture both undamaged and damaged buildings in contrast to the pre-event images. Moreover, images immediately after the disaster are usually of lower quality, as there is a less chance to obtain good projection with proper atmospheric conditions in the short period of time.  Therefore, the performance is lower on them compared to the pre-event images, where the best ones are chosen from a sequence of images over the longer period of time.

  \begin{table}
      \centering
      \caption{Prediction quality of bounding boxes for building footprint detection. Values are given in the percentage form.}\label{table:prediction_quality_of_all_bounding_boxes}
      \begin{tabular}{r l l l l}
        \toprule
          Timestamp & AP\textsuperscript{50} & F\textsubscript{1} & Precision & Recall\\ 
        \midrule
          Pre-event & 38 & 57 & 52 & 63 \\
          Post-event & 23 & 47 & 44 & 49 \\ 
        \bottomrule
      \end{tabular}

  \end{table}

\subsection{Mark Aggregation}\label{sec:dataset:marks_aggregation}
  The crowd was asked to provide marks of three possible types that correspond to the severity of structural damage: \textit{minor}, \textit{significant}, and \textit{catastrophic} damage. The details are given in \figurename~\ref{figure:severity_examples} in the appendix.  Each mark represents a point on the map with given geospatial coordinates of latitude and longitude and label of the class.
 
  The footprint detection step yields the buildings' bounding boxes.  For each object, there is a set of corresponding mark labels from volunteers.  Due to volunteers being imperfect, the marks are noisy, meaning that labels from different volunteers for the same objects may identify different levels of damage severity.  In the label aggregation step we determine for each object the consensus label from the crowd labels. The standard input for such a crowdsourcing task is an object/volunteer matrix, where each cell contains a label from the corresponding volunteer for this particular object.  To form this matrix in our case, in addition to the explicit labels from volunteers (i.e., \textit{minor}, \textit{significant}, and \textit{catastrophic} damage in our case), we use {\it unseen} and {\it empty} labels.  An {\it unseen} label indicates that this particular volunteer has not seen the image that contains this object.  Almost every volunteer will have some unseen labels, as it is rare that a single volunteer labels all images over an entire dataset.  {\it Empty} are objects where a volunteer has not marked damage, either by labeling the whole image as undamaged, or labeling the damage in other parts of the image only.

  There are multiple approaches for crowdsourcing aggregation that can be used once the object/volunteer matrix has been constructed.  We consider below the \emph{majority voting} (MV) and \emph{independent Bayesian classifier combination} (IBCC)  models.
\subsubsection{Majority Voting}\label{sec:dataset:marks_aggregation:majority_voting}
  The majority voting model treats all volunteer marks to be of equal importance.  For each object the most common label is selected.  When the data quality is low, a significant number of volunteers can miss true objects and there will be mostly empty labels. Therefore, we weighted the labels, with a lower value for the empty label.  However, volunteers differ in skill and some volunteers may identify the damage better than others, but majority voting will treat their labels with the same importance.  This can partially be solved by weighting the different volunteers according to their skill level.  This insight is incorporated into the independent Bayesian classifier combination approach.

\subsubsection{Independent Bayesian Classifier Combination}\label{sec:dataset:marks_aggregation:independent_bayesian_classifier_combination}
The Dawid-Skene model~\citep{dawid1979maximum} introduces weights for the quality of labels of each volunteer through a confusion matrix.  Each row of the confusion matrix is the probability a volunteer assigns a label to an object conditional on the true class of the object.  The individual confusion matrices for each volunteer are learnt from the data and the labels are subsequently aggregated. 

The Bayesian version of the Dawid-Skene model, called independent Bayesian classifier combination~\citep{kim2012bayesian}, places a Dirichlet prior over class probabilities and also over each row of the confusion matrix for each volunteer.  These priors can express the skill level of the volunteer when known. Approximations for the posterior object class probabilities and the posterior distributions over the confusion matrices are calculated efficiently using variational Bayes~\citep{simpson2013dynamic,isupova2018bccnet}.

To compare the accuracy of IBCC and MV we used the expertly annotated data as ground truth.  The results are given in Table~\ref{table:classification_accuracy_of_aggregation_models}.  It is not possible to choose the empty label weight such that MV aggregates several noisy points into the correct empty label, as the optimal weight varies in different cases. Therefore, it leads to overestimation of damaged buildings by MV.  This is the main reason why IBCC shows better performance according to the metrics.  Examples of the results of the different models are shown in \figurename~\ref{figure:aggregation_comparison} in the appendix.

    \begin{table}
      \centering
      \caption{Classification accuracy of mark aggregation models. Values are F1 scores in the percentage form. Column names contain the support size for the metric. Average values for all labels are weighted by support. }\label{table:classification_accuracy_of_aggregation_models}
      \begin{tabular}{r l l l l l}
        \toprule
          Model & average / 135 & empty / 80 & minor / 12 & significant / 30 & catastrophic / 13 \\ 
        \midrule
          IBCC & 92 & 92 & 40 & 58 & 65 \\
          MV & 90 & 70 & 40 & 60 & 59 \\ 
        \bottomrule
      \end{tabular}
 
  \end{table}

\section{Damage Detection}\label{sec:damage_detection}
  In this section we demonstrate the viability of our approach for building a training dataset for damage detection from point crowdsourced marks by training a neural net on this dataset.  In this case we train using post-event images only.  
  
  For the neural network architecture we use Faster-RCNN \citep{ren2015faster} with ResNet-50+FPN backbone implemented in detectron2 library~(\url{github.com/facebookresearch/detectron2}) pre-trained on ImageNet.  We achieve the results on the COCO metrics for the bounding boxes presented in Table~\ref{table:bbox_prediction_quality}.  For the reference, this model achieves AP of 37.9 on the COCO2017 dataset~(\url{github.com/facebookresearch/detectron2/blob/main/MODEL_ZOO.md}).  
  
 \begin{table}
      \centering
      \caption{Bounding boxes with damage severity prediction quality on the test data. }\label{table:bbox_prediction_quality}
      \begin{tabular}{l l l l l l}
        \toprule
          AP & AP50 & AP75 & APs & APm & APl\\ 
        \midrule
          18.488 & 31.169 & 21.876 & 16.170 & 27.217 & 0.000 \\
        \bottomrule
      \end{tabular}
 
  \end{table}
  
\begin{figure}
    \centering
    \begin{subfigure}[b]{0.49\textwidth}
        \includegraphics[width=\textwidth]{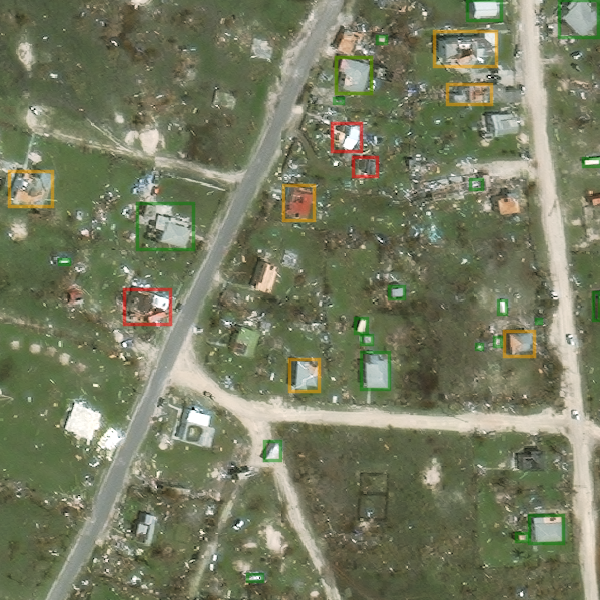}
        \caption{Predictions}\label{figure:damage_detection:barbuda_2545_pred}
    \end{subfigure}
    \begin{subfigure}[b]{0.49\textwidth}
        \includegraphics[width=\textwidth]{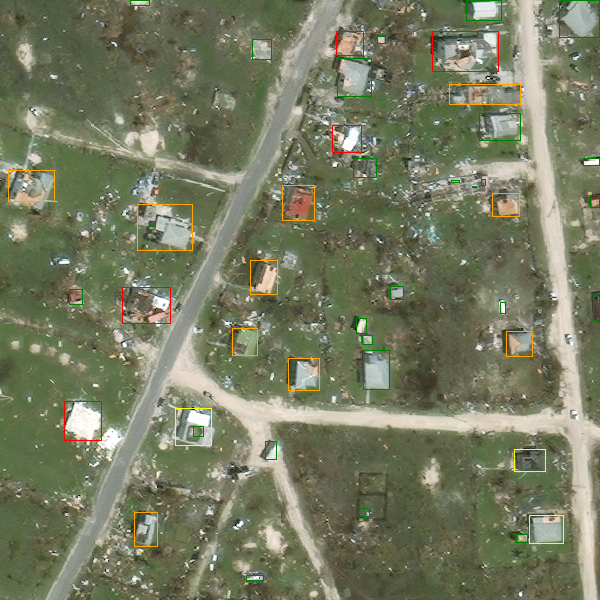}
        \caption{Ground truth}\label{figure:damage_detection:barbuda_2545_gt}
    \end{subfigure}
    \caption{Damage detection by the neural network. Ground truth is generated with the approach described in Section~\ref{sec:dataset}. Colors of bounding boxes represent labels: green~---~empty, i.e., undamaged, yellow~---~minor, orange~---~significant, and red~---~catastrophic. }\label{figure:damage_detection_main}
  \end{figure}

  \figurename~\ref{figure:damage_detection_main} shows example predictions on the test data, additional examples are provided in the appendix. Compared to the COCO dataset, in our case the objects are generally smaller (for example, we have no objects in \textit{large} category of bounding boxes) and the image quality is reduced due to the environment. Moreover, the size of our training dataset is much smaller than the COCO dataset size and training labels in the COCO dataset are carefully tuned. Nevertheless, as it can be seen in \figurename~\ref{figure:damage_detection_main} the trained neural network is able to correctly identify approximate areas of damage. This is the information actually required by the disaster responders as they use damage mapping to plan their operation and logistic. Potential errors in terms of individual buildings are not very crucial for this purpose.
  
\section{Conclusions}\label{sec:conclusion}
  We propose a crowdsourced point-based labelling strategy that reduces the time to create structural damage datasets following disasters. Our method can be used to classify the damage severity of individual buildings in the disaster zone during the live response operations. We rely on only point marks placed anywhere on damaged structural buildings. This allows us to employ non-expert volunteers and to speed up the labelling process. The approach is robust to labelling inaccuracies of different volunteers and to common misalignments of pre-~and post-event satellite imagery.   We have demonstrated that these datasets can be used to train a neural network object detection damage mapper with the sound accuracy. This mapper can then be used to label new areas affected by the disaster during the same response deployment.  

\section{Acknowledgements}
DK and BDS acknowledge support from UKRI via the Medical Research Council [FLF grant number MR/T044136/1] and the Biotechnology and Biological Sciences Research Council [grant number BB/T018941/1]. DK and BDS also acknowledge Lancaster University IAA funding support.
SR acknowledges support from the Lloyd's Register Foundation through the Alan Turing Institute’s Data Centric Engineering programme and for EPSRC funding as a Researcher in Residence with the Satellite Applications Catapult. DK, BDS and SR acknowledge support from the Science and Technology Facilities Council [grant number ST/S00307X/1].

{\small
\bibliography{bibliography}
}

\newpage
\section{Appendix}

In the appendix we provide additional figures to the main text. 

\begin{figure}
    \centering
    \begin{subfigure}[b]{0.32\textwidth}
        \includegraphics[width=\textwidth]{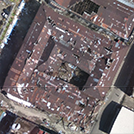}
        \caption{Minor}\label{figure:severity_examples:minor}
    \end{subfigure}
    \begin{subfigure}[b]{0.32\textwidth}
        \includegraphics[width=\textwidth]{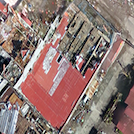}
        \caption{Significant}\label{figure:severity_examples:significant}
    \end{subfigure}
    \begin{subfigure}[b]{0.32\textwidth}
        \includegraphics[width=\textwidth]{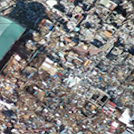}
        \caption{Catastrophic}\label{figure:severity_examples:catastrophic}
    \end{subfigure}
    \caption{Severity of structural damage. \textit{Minor}~(\protect\subref{figure:severity_examples:minor}) (less than 20\% of the building is damaged) where there is clearly damage, but the structure seems like it could still be used or inhabited; \textit{Significant}~(\protect\subref{figure:severity_examples:significant}) (between 20\% and 60\%) where the damage is severe, but the structure is still present and appears recognizable, even if it may be too damaged to be used or inhabited and
    \textit{Catastrophic}~(\protect\subref{figure:severity_examples:catastrophic}) (over 60\%) for which the structure is so badly damaged it is clearly unusable, and may not even exist anymore.}\label{figure:severity_examples}
    \end{figure}

\figurename~\ref{figure:severity_examples} provides examples for each of the damage severity classes that we use in this work.

\begin{figure}
    \centering
    \begin{subfigure}[b]{0.49\textwidth}
        \includegraphics[width=\textwidth]{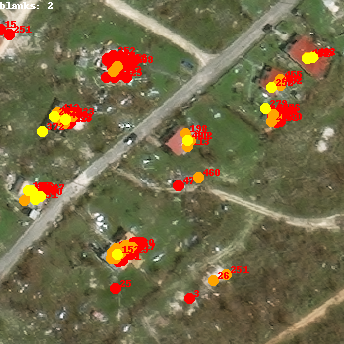}
        \caption{Point marks}\label{figure:aggregation_comparison:marks}
    \end{subfigure}
    \begin{subfigure}[b]{0.49\textwidth}
        \includegraphics[width=\textwidth]{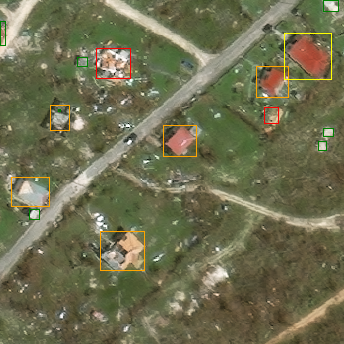}
        \caption{MV aggregation}\label{figure:aggregation_comparison:mv}
    \end{subfigure}
    \begin{subfigure}[b]{0.49\textwidth}
        \includegraphics[width=\textwidth]{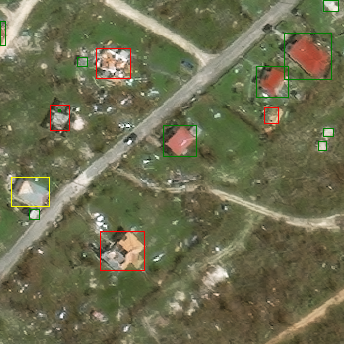}
        \caption{IBCC aggregation}\label{figure:aggregation_comparison:ibcc}
    \end{subfigure}
    \caption{Comparison of crowdsourcing aggregation models. Marks from different volunteers~(\protect\subref{figure:aggregation_comparison:marks}) are aggregated using the MV~(\protect\subref{figure:aggregation_comparison:mv}) and IBCC~(\protect\subref{figure:aggregation_comparison:ibcc}) algorithms inside the detected building footprints. Colours indicate severity of damage: yellow~---~minor, orange~---~significant, red~---~catastrophic.  On the image~\protect\subref{figure:aggregation_comparison:marks} the number of volunteers marked this image as empty of damage is depicted in the top left; also the image contains IDs of the volunteers that placed each damage mark.}\label{figure:aggregation_comparison}
  \end{figure}

Illustration of mark aggregation step with majority voting and independent Bayesian classifier combination (Section~\ref{sec:dataset:marks_aggregation}) is given in \figurename~\ref{figure:aggregation_comparison}.    
    
In addition to \figurename~\ref{figure:damage_detection_main}, we provide further examples of damage detection by the trained neural network (Section~\ref{sec:damage_detection}) on the test data.
  
  \begin{figure}
    \centering
    \begin{subfigure}[b]{0.49\textwidth}
        \includegraphics[width=\textwidth]{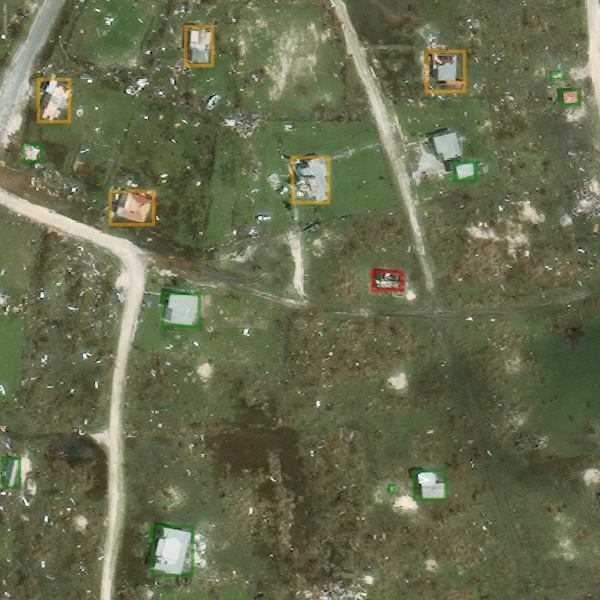}
        \caption{Prediction 1}\label{figure:aggregation_comparison:barbuda_2608_pred}
    \end{subfigure}
    \begin{subfigure}[b]{0.49\textwidth}
        \includegraphics[width=\textwidth]{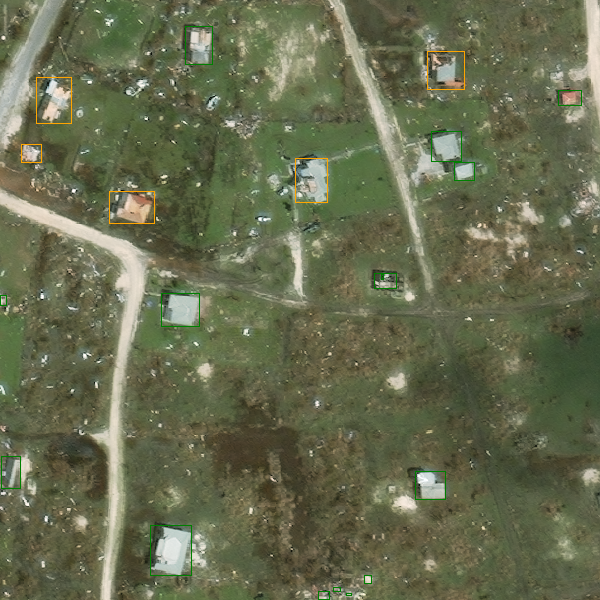}
        \caption{Ground truth 1}\label{figure:aggregation_comparison:barbuda_2608_gt}
    \end{subfigure}\\
    \begin{subfigure}[b]{0.49\textwidth}
        \includegraphics[width=\textwidth]{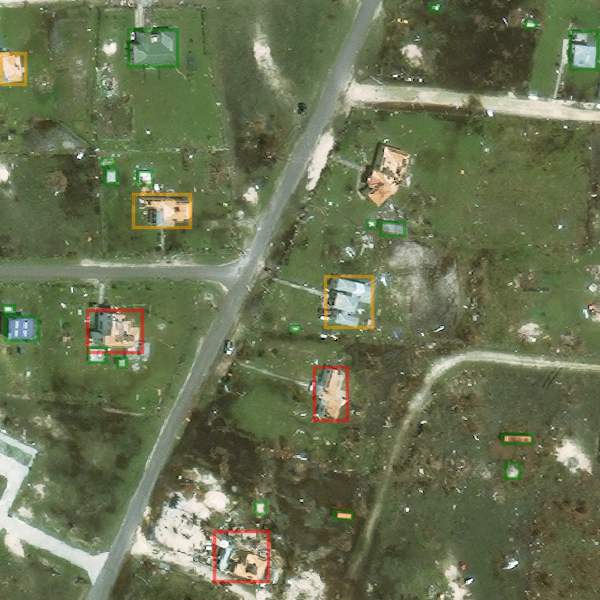}
        \caption{Prediction 2}\label{figure:aggregation_comparison:barbuda_2670_pred}
    \end{subfigure}
    \begin{subfigure}[b]{0.49\textwidth}
        \includegraphics[width=\textwidth]{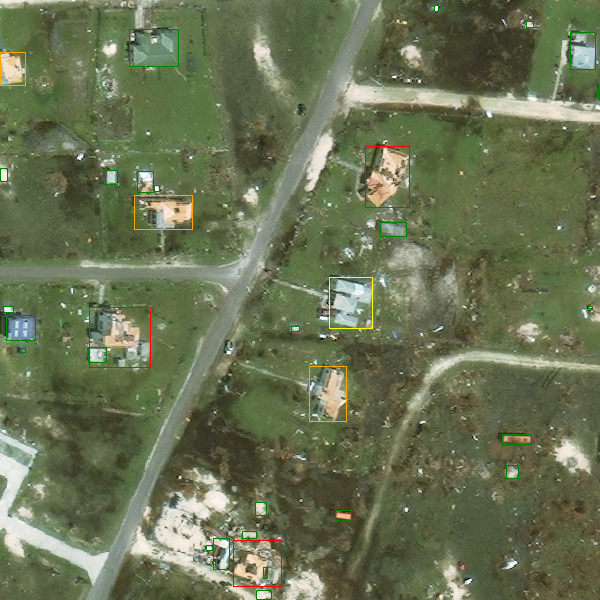}
        \caption{Ground truth 2}\label{figure:aggregation_comparison:barbuda_2670_gt}
    \end{subfigure}
    \caption{Damage detection examples}\label{figure:damage_detection}
  \end{figure}
\end{document}